\documentclass[runningheads]{llncs}
\usepackage{graphicx}
\usepackage{lscape}      
\usepackage{longtable}   
\usepackage{afterpage}
\usepackage{breakurl}
\begin{document}
\title{Experts-in-the-Loop: Establishing an Effective Workflow in Crafting Privacy Q\&A}

\author{Zahra Kolagar\orcidID{0000-0002-3917-5655} \and
Anna Katharina Leschanowsky\orcidID{0000-0003-2994-2336} \and
Birgit Popp\orcidID{0000-0003-2193-9861}}
\titlerunning{Experts-in-the-Loop: Crafting Privacy Q\&A}
\authorrunning{Z. Kolagar et al.}
%
\institute{Fraunhofer IIS, Erlangen, Germany\and
\email{\{zahra.kolagar,anna.leschanowsky,birgit.popp\}@iis.fraunhofer.de}}

\maketitle              
\begin{abstract}

Privacy policies play a vital role in safeguarding user privacy as legal jurisdictions worldwide emphasize the need for transparent data processing. While the suitability of privacy policies to enhance transparency has been critically discussed, employing conversational AI systems presents unique challenges in informing users effectively. In this position paper, we propose a dynamic workflow for transforming privacy policies into privacy question-and-answer (Q\&A) pairs to make privacy policies easily accessible through conversational AI. 
Thereby, we facilitate interdisciplinary collaboration among legal experts and conversation designers, while also considering the utilization of large language models' generative capabilities and addressing associated challenges. Our proposed workflow underscores continuous improvement and monitoring throughout the construction of privacy Q\&As, advocating for comprehensive review and refinement through an experts-in-the-loop approach.

\end{abstract}

\section{Introduction}

Legal jurisdictions worldwide ask providers to make their privacy policies accessible and easily understandable for users~\cite{GDPR,CCPA}. In the European Union, the General Data Protection Regulation (GDPR) has established the ``Right to be informed" while in the US most jurisdictions rely on the ``Notice and Choice" principle~\cite{GDPR,ProtectingConsumerPrivacy2012}.

However, privacy policies are only rarely read and are not suitable for promoting transparency~\cite{mcdonald2008cost,cateLimitsNoticeChoice2010,cranorNecessaryNotSufficient2012,schaubDesignSpaceEffective2018}. This problem is exacerbated in interaction with conversational AI systems as they use natural language to communicate with users and often require unfavorable modality switching to sufficiently inform users. Conversational privacy systems or privacy Q\&A can be a means to deliver privacy policies interactively~\cite{harkousPriBotsConversationalPrivacy,pearman2022user}. 
Previous research on improving consent flows within conversational AI interfaces has shown that traditional approaches are not sufficient for appropriately informing users and has advocated for conversational approaches~\cite{pearman2022user}. 

Yet, technical and legal challenges have so far hindered the widespread adoption of privacy question-answering (see Table~\ref{tab:ConversationalPrivacyExamples} for examples of privacy Q\&A in current systems). 
One stream of research has applied natural language processing (NLP) to analyze and extract meaningful information from privacy policies and overcome technical difficulties by introducing various annotation schemes and collecting privacy policy corpora~\cite{ahmadPolicyQAReadingComprehension2020a,wilsonCrowdsourcingAnnotationsWebsites2016,wilson2016creation}. However, current privacy Q\&A systems rely on the original privacy policy texts for constructing privacy answers regardless of the known readability and understandability issues of legal texts~\cite{ravichanderBreakingWallsText2021}.
Given the increasing prominence of Large Language Models (LLMs), which have found applications in question-answering tasks~\cite{zhang2022greaselm} or have been incorporated into conversational AI systems through various prompting techniques~\cite{chen2023llm,semnani2023wikichat,lee2023prompted}, it is conceivable that companies will leverage LLM capabilities in the context of privacy Q\&A. Nevertheless, legality and usability challenges remain due to their lack of fact-checking and groundedness capabilities~\cite{KASNECI2023102274,peng2023check,ray2023chatgpt,teubner2023welcome,thoppilan2022lamda} and the potentially negative impact of automatically generated answers on user experience. While considering that privacy policies are legally binding documents, it is yet unclear how automatically generated privacy answers can ensure legal correctness and user-friendliness. 
Therefore, in this position paper, we propose an interdisciplinary workflow to address the shortcomings of current privacy question-answer systems and to ensure both legal validity and usability. We address the following questions: 

\begin{itemize}
    \item RQ 1: What are the essential components of an effective workflow for privacy Q\&A?
    \item RQ 2: How does an interdisciplinary collaboration between legal experts and conversation designers form within this workflow?
\end{itemize}


\section{Proposed Workflow} \label{sec:proposedWorkflow}

\begin{figure*}[ht]
  \includegraphics[width=\linewidth]{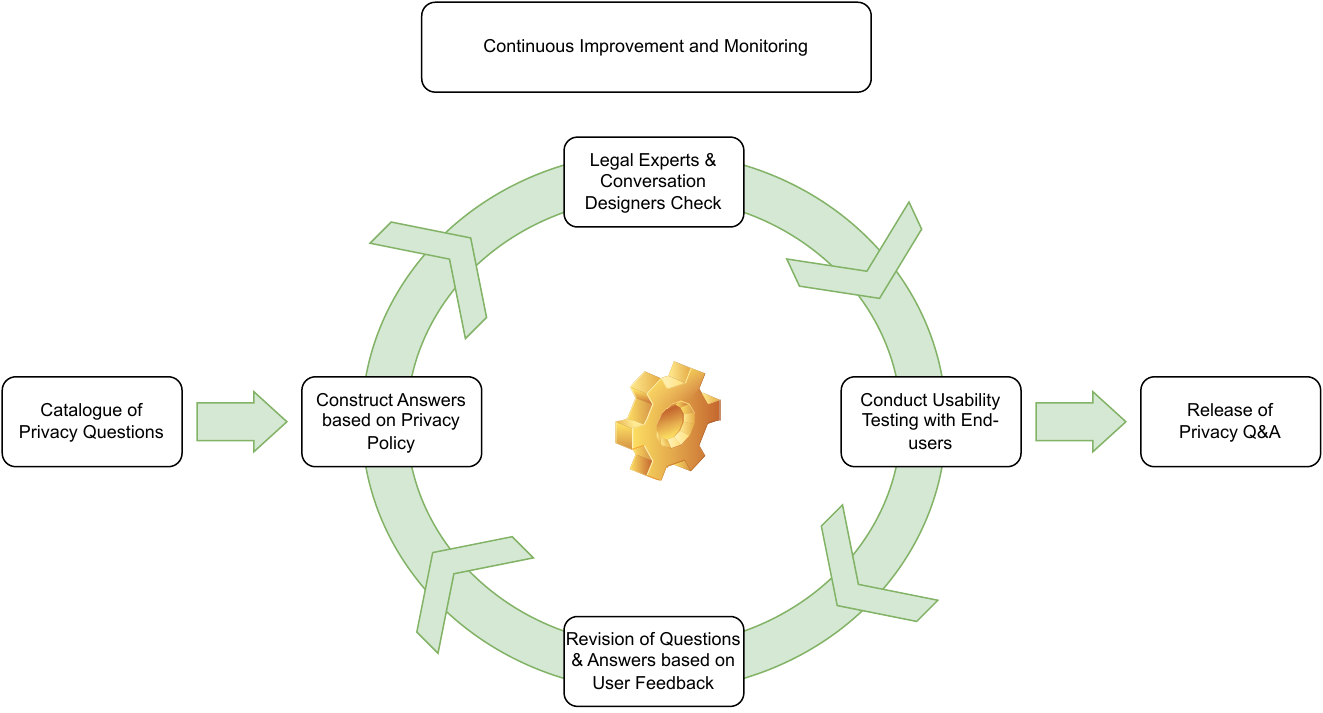}
  \caption{workflow for constructing privacy Q\&As}
 \label{fig:privacyqa}
\end{figure*}

Figure~\ref{fig:privacyqa} shows our proposed workflow for constructing privacy Q\&A for conversational AI systems and emphasizes the importance of continuous improvement and monitoring. 


\subsection{Catalogue of Privacy Questions} \label{Catalogue of Privacy Questions}

Initially, a comprehensive set of privacy questions needs to be gathered which constitutes the backbone of privacy Q\&A. It allows users to easily access information without the need to read lengthy and hardly understandable privacy policies~\cite{mcdonald2008cost}. The catalogue of questions can be derived from existing privacy Q\&A corpora~\cite{ahmadPolicyQAReadingComprehension2020a,ravichander-etal-2019-question} or from user engagement and studies. While existing corpora have not yet included conversational AI system-specific questions, e.g. ``Can I delete my voice recordings?", they can serve as a suitable starting point. In addition, user studies based on scenario-driven surveys; so-called vignette surveys~\cite{atzmuller2010experimental}, can produce more specific questions. By designing scenarios and asking participants to draw upon their personal experiences, one can encourage them to generate relevant questions for various topics. This ``human-in-the-loop" approach allows for a more comprehensive and user-centric creation of privacy questions. 

Drawing from the findings of Ahmad et al. ~\cite{ahmadPolicyQAReadingComprehension2020a}, Wilson et al. ~\cite{wilson2016creation} and Ravichander et al.~\cite{ravichander-etal-2019-question}, we recommend incorporating at least the following themes into the collection of privacy questions to address users' concerns about privacy policies and the management of personal data: ``First Party Collection/Use", ``Third Party Sharing/Collection", ``Data Security", ``Data Retention", ``User Access, Edit and Deletion", and ``User Choice/Control". Additional categories may arise during the iterative process and can be added as users engage with the privacy Q\&A.

To improve user accessibility and prevent misunderstanding of the term ``data", we additionally propose organizing the collection of privacy questions into distinct topics including "data collected about you" (e.g., name, address, time zone, and age), "data collected about your contacts" (e.g., names of stored contacts), "data collected about your files and activities," and "data collected about your device and network" (e.g., WiFi information)~\cite{Cohen_2022}. Such categorization will facilitate users' ability to search for and locate specific information more efficiently. 

Relying on multiple sources to collect privacy questions can lead to a growing catalogue, making comprehensive reviews time-consuming and expensive. While reviews can help to reduce the size of the catalogue, e.g. by deleting duplicates, multiple questions can be matched with the same or similar answers minimizing the workload for human experts in subsequent steps. Semantic Textual Similarity (STS) approaches such as Sentence-BERT can be used for the matching to identify the most representative question per category and topic~\cite{reimers19sentence}, However, this approach does not exclude the emergence of new categories and topics.

In general, we recommend commencing with a compact set of questions that encompasses all crucial categories and topics and revising or expanding on the catalogue iteratively. The determination of what is deemed essential and the number of questions needed to train a language model can vary, influenced by factors including the specific privacy policy, use cases, business models, and, most notably, the chosen language model and training approach.

\subsection{Construction of Answers based on Privacy Policy} \label{Construction of Answers based on Privacy Policy}

After privacy questions are categorized and reviewed, relevant information from the privacy policies must be extracted in the form of a sentence selection extraction task~\cite{ravichander-etal-2019-question} and translated into concise answers without legal jargon, prioritizing simplicity, transparency, and usability requirements.

While recent NLP approaches can generate direct answers to user queries using guided text summarization, reading comprehension, or a combination of both, their primary focus is not on privacy Q\&A tasks~\cite{widyassari2022review,kryscinski2019neural,keymanesh2021fairness,sarkhel2020interpretable,keymanesh2021privacy}.
Yet, one of the hindrances of widespread adaptation of privacy Q\&As is the amount of accurately selected and annotated data needed for training a language model for privacy Q\&A~\cite{ravichander-etal-2019-question}. While augmentation techniques can help overcome those challenges~\cite{parvez2022retrieval}, they may introduce potential biases or inaccuracies. Moreover, few-shot learning approaches, which rely on a smaller amount of data, struggle to handle the complexities inherent in privacy policies as they encompass nuanced language and legal terminology that require a comprehensive understanding of the context~\cite{song2022comprehensive}. 
Few-shot models may also encounter difficulties in comprehending intricate details and generating accurate answers without extensive training. Finally, using LLMs to create privacy Q\&A can result in plausible yet potentially incorrect or misleading information~\cite{weidinger2022taxonomy}. While these advances can reduce workload, increase the efficiency of the workflow, and provide initial privacy answers, due to current limitations they cannot ensure legally valid and user-friendly privacy Q\&As. Therefore inspired by well-established ``human-in-the-loop" processes, we propose an ``experts-in-the-loop" approach as discussed in-depth in the next section.

\subsection{Legal Experts and Conversation Designers Check}

Involving experts is crucial to ensure the legality and usability of privacy Q\&A and mitigate the risks of NLP approaches. 
While legal experts can validate generated answers on legal correctness, they can also provide valuable input in crafting prompts to harness the capabilities of LLMs~\cite{ge2023openagi,zamfirescu2023johnny}. Well-designed prompts that explicitly specify the context of privacy-related questions can guide LLMs to generate more accurate and relevant answers~\cite{white2023prompt,liu2023pre}. 
Above that, users often ask out-of-scope or overly specific questions that vary in style and language from privacy policies, posing challenges for automated assistants to grasp the intent of the user and find relevant information \cite{ravichander-etal-2019-question}. 
While LLMs may be able to answer in a compelling manner to out-of-scope questions, their answers may be inconsistent, not legally valid, and subject to hallucinations~\cite{weidinger2022taxonomy,beutel2023artificial}. Therefore, legal experts play a crucial role in validating the answers by assessing their lawfulness, accuracy, and compliance with data protection regulations.

In addition, answers based on NLP approaches may not follow best practices of conversation design~\cite{harkous2018polisis}. They might include long answers due to the nature of privacy policies and thereby violate the conversation design principle of minimization~\cite{MooreNCF}. Even if relying on LLMs for constructing answers, they may tend to generate lengthy and less concrete responses as opposed to human experts (see ChatGPT example responses in Table~\ref{tab:ConversationalPrivacyExamples})~\cite{guo2023close}. Moreover, they may incorporate technical and legal terms that are difficult for users to understand and may require additional repair and rephrasing techniques~\cite{MooreNCF}. While usability testing can provide insights into user experience and satisfaction with privacy Q\&A, it cannot replace the expertise of conversation designers in crafting, fine-tuning, and repairing users' conversations with conversational AI systems. Therefore, we emphasize the role of conversation designers in the proposed workflow to identify violations of design principals~\cite{MooreNCF} and ensure clarity, simplicity, and comprehensibility of the privacy-related answers.   

Overall, the involvement of experts should be strategic and timely. Legal experts can play a more proactive role in reviewing and validating a subset of responses or questions, with a focus on those flagged as critical or potentially non-compliant. Additionally, conversation designers' expertise is instrumental in identifying design violations and ensuring the clarity, simplicity, and comprehensibility of privacy-related answers. This approach ensures that the expert's involvement is targeted and efficient.

\subsection{Usability Testing \& Revision}

Usability testing is crucial to evaluate privacy Q\&A in real-world scenarios and uncover issues not caught by human experts. This could be achieved by presenting participants with realistic privacy scenarios~\cite{habib2020s}, allowing them to ask privacy-related questions and receive associated answers from the conversational AI system. Feedback from the participants can include ``user needs", ``user ability and effort", ``user awareness", ``user comprehension", ``user sentiment", and ``decision reversal"~\cite{habib2022evaluating}.

Finally, the integration of user feedback as shown in Figure~\ref{fig:privacyqa} is essential to enhance the development of lawful and user-friendly privacy Q\&As. 
In addition to privacy policy changes, the feedback and study results should serve as triggers for subsequent rounds of collecting and refining privacy questions and answers and gathering experts' opinions. Despite the potential cost and time investment, this iterative process guarantees the inclusion of users' inquiries, ensures that answers are both lawful and user-friendly and validates the iterative approach through empirical research.

\section{Discussion and Future Work}
\label{sec:DiscussionandFutureWork}

While legal regulations emphasize the need to appropriately inform users, privacy policies have long failed to enhance transparency -- an issue that is exacerbated by conversational AI systems as they use natural language to exchange information with users. 
Despite recent advancements, privacy Q\&A has not yet reached widespread adoption due to the limited number of available datasets and technical challenges to ensure legal correctness and user-friendliness. Our proposed workflow can be used to effectively create understandable and legally accurate privacy Q\&A that enables information on data processing practices to be easily accessible in conversational AI systems. This workflow can be used to collect datasets on privacy question-answering to train future language technology and enable broader implementation of privacy Q\&A. For creating such corpora and meeting end-users' needs such as helping them make informed decisions and allowing them to have control over their personal data interactively, interdisciplinary work of legal experts, engineers, and dialogue designers is required.

In the future, we aim to test the workflow in practice by collecting a privacy Q\&A dataset evaluated and revised by a multidisciplinary team of experts and conducting usability testing. 
This will not only allow us to show the feasibility of our proposed workflow but to introduce a novel privacy Q\&A dataset to the research community that ensures legal accuracy and usability and helps to tackle long-lasting problems in the widespread adoption of privacy Q\&A.  

We believe that our workflow is designed to withstand the test of time, even with the emergence of new approaches, such as companies incorporating LLMs capabilities to create privacy Q\&As in the future. By leveraging the expertise of both legal professionals and conversation designers, we can enhance and fine-tune the performance of LLMs through the prompting and inclusion of conversation design principles. Experts' opinions are essential for designing well-crafted prompts that explicitly define the context of privacy-related questions. Thereby, incorporating legal terminology and knowledge of privacy and conversation design patterns~\cite{MooreNCF,lenhardLiteratureStudyPrivacy2017} can guide the model in providing appropriate responses that are both legally acceptable and user-friendly. Therefore, future work could research prompts for privacy Q\&A and evaluate the impact of existing design patterns and experts' opinions.

\section{Conclusion}

In conclusion, our research presents a workflow that guides the transformation from a collection of privacy questions and policies to privacy question-and-answer pairs. This workflow facilitates the integration of legal experts, conversation designers, and user feedback, ultimately leading to the release of privacy Q\&As to the public. We believe the workflow will be valuable for researchers and organizations in their efforts to provide lawful and user-friendly privacy Q\&As.

\section*{Limitations}

Firstly, we have not yet evaluated the workflow in real-life scenarios and usability testing will be subject to future research. Moreover, we relied on a limited number of legal experts and conversation designers for the construction of the workflow, and input from a wider range of experts is needed for further improvement. On the other hand, limitations might arise when incorporating legal experts and conversation designers into the workflow. Timely availability, cost management, and reconciling varying perspectives between legal compliance and conversational design pose notable constraints. As the volume of user queries grows, maintaining scalability, consistency, and coordination among experts becomes more demanding. These limitations call for further exploration in future research.  

Finally, while we believe that our proposed workflow is applicable to conversational AI systems in general, specific applications might require adaptations or additional experts to be included in the process, e.g. medical applications might require medical expertise in addition to conversation designers. 

\section*{Ethics Statement}

This research does not involve any discrimination, social bias, or private data, and all the examples are listed in the Appendix.

\section*{Acknowledgements}

We extend our heartfelt gratitude to the anonymous reviewers for their valuable insights.

\bibliographystyle{splncs04}
\bibliography{paper}

\appendix
\section*{Appendix}

\small
\begin{landscape}
\begin{longtable}{|p{0.08\linewidth}|p{0.08\linewidth}|p{0.15\linewidth}|p{0.15\linewidth}|p{0.1\linewidth}|p{0.4\linewidth}|}
\caption{Examples of Privacy Q\&A of common CAI systems. Example Questions are taken from~\cite{ahmadPolicyQAReadingComprehension2020a} and adapted to fit the context of CAI. Answers were derived on the 15th of June 2023 in Germany, and a correct understanding of questions was ensured in the case of the voice assistants.}
\label{tab:ConversationalPrivacyExamples}
\endfoot
\hline
\textbf{Category} & \textbf{Question} & \multicolumn{4}{c|}{\textbf{CAI System's Answer}} \\
\hline

& & \textbf{Amazon Alexa} & \textbf{Google Assistant} & \textbf{Apple Siri} & \textbf{ChatGPT} \\
\hline
\endhead

First Party Collection/ Use & Why do you collect my data? & Your data allows me to respond to you and helps me learn from our interactions so that I can better help you. For example, the more variation in spoken language I learn, the better I'll be at understanding what you mean. You can find settings for how your data is used by visiting \burl{amazon.co.uk/ alexaprivacysettings} or the Privacy section of the Alexa App. & Google collects data when you provide it to them, and when you use Google services like Search or Maps. To learn more, visit Google's Safety Center. & I don't understand 'Hey Siri why do you collect my data? ', but I could search the web for it. & As an AI language model, I don't have the ability to collect or store personal data from individuals unless it has been shared with me during our conversation. I am designed to respect user privacy and confidentiality. However, it's important to note that as an AI developed by OpenAI, they may collect and store user data as described in their privacy policy. The purpose and extent of data collection, if any, would be outlined by OpenAI in their privacy policy, which you can review for more specific information.\\\hline
Third-Party Sharing/ Collection & Do you share my information with others? & Amazon takes privacy seriously. For more information and to view Amazon's privacy notice, visit the help section of your Alexa app. & Google collects data when you provide it to them, and when you use Google services like Search or Maps. To learn more, visit Google's Safety Center. & Hmm... I don't have an answer for that. Is there something else I can help with? & As an AI language model, I don't have access to personal data about individuals unless it has been shared with me during our conversation. I am designed to respect user privacy and confidentiality. I can't share or distribute any personal information because I don't have the capability to store such information in the first place. However, it's important to note that as an AI developed by OpenAI, they may collect and store user data as described in their privacy policy. The specifics of data sharing, if any, would be outlined in OpenAI's privacy policy. I recommend reviewing their privacy policy for more details on how they handle and protect user information. \\\hline
Data Security & Do you use encryption to secure my data? & \textit{No response} & \textit{Refers to a Website} & Here's what I have found. \textit{Refers to some Websites}& As an AI language model, I don't process or store personal data about individuals unless explicitly provided to me during the course of our conversation. 
I don't have access to the infrastructure or storage systems to encrypt or secure any data. However, it's important to note that data security practices and encryption methods, if any, would be implemented by the organization or platform hosting and operating me. For example, OpenAI, the organization behind ChatGPT, may have security measures in place to protect user data according to their privacy policy and industry best practices. I recommend reviewing OpenAI's privacy policy or reaching out to them directly for specific details on how they handle data security and encryption. \\\hline
Data Retention & How long will you keep my data? & I don't know that one. & \textit{Refers to a Website} & I have found this on web. \textit{Refers to some Websites} & As an AI language model, I don't have the ability to store personal data or retain information about individuals unless explicitly provided to me in the course of our conversation. I am designed to respect user privacy and confidentiality. My primary function is to provide information and answer questions to the best of my knowledge and abilities in the present moment. Once the conversation ends, the data is not stored for future reference. However, it's important to note that the policies regarding data retention, if any, would be determined by the organization or platform that hosts and operates me. For specific information on data retention, I recommend reviewing the privacy policy of the organization or platform you are interacting with, such as OpenAI in the case of ChatGPT. \\\hline
User Access, Edit and Deletion & Will you let me access and edit my data? & \textit{Refers to a Website} & \textit{Toggling Mobile Data} & Sorry, I can't do that. & As an AI language model, I don't have the capability to provide access to or allow editing of personal data. I don't store personal data unless explicitly provided to me during our conversation, and once the conversation ends, the data is not retained. If you have provided personal data to a platform or service that utilizes AI models like me, such as OpenAI's ChatGPT, you would need to refer to the platform's privacy policy and data management practices to understand how to access and edit your data. They would be responsible for handling and managing user data and would have procedures in place to address such requests. \\\hline
User Choice/ Control & What use of information does the user choice apply to? & I don't know that one. & \textit{Refers to a Website} & I have found this on web. \textit{Refers to some Websites} & When referring to the "user choice," it is important to clarify the specific context or platform to which it applies. The concept of user choice generally relates to the control users have over the collection, use, and sharing of their personal information. 

In the context of an AI platform or service, such as OpenAI's ChatGPT, the user choice would typically apply to how the platform handles and utilizes the user's personal data. This can include aspects such as:
\begin{enumerate}
    \item Data collection: Users may have the choice to provide certain personal information voluntarily or to limit the amount of data collected by the platform.
    \item Data usage: Users may have the choice to specify how their personal information is used by the platform, such as for improving the service, customizing recommendations, or for research purposes.
    \item Data sharing: Users may have the choice to determine whether their personal information is shared with third parties or used for targeted advertising.
    \item Data retention and deletion: Users may have the choice to request access to their personal data, request modifications or corrections, or even request the deletion of their data from the platform's records.
\end{enumerate} 

However, it's important to note that the specific user choices available may vary depending on the platform, service, or organization in question. To understand the scope and details of user choice for a particular platform, it is advisable to review their privacy policy, terms of service, or directly contact the platform provider for clarification.\\
\hline

\end{longtable}
\end{landscape}

\end{document}